\documentclass[12pt]{elsarticle}

\usepackage[english]{babel}
\usepackage{amsfonts}
\usepackage[utf8]{inputenc}
\usepackage{natbib}
\usepackage{xspace}
\usepackage{graphicx}
\usepackage{todonotes}
\usepackage{pythonhighlight}
\usepackage[ruled, linesnumbered]{algorithm2e}
\usepackage{pgf-umlcd}
\usepackage[htt]{hyphenat}
\usepackage{listings}
\usepackage{xcolor}
\usepackage{placeins}
\usepackage{hyperref}

\setcitestyle{square, comma, numbers, sort&compress}
\journal{ }
\usepackage{etoolbox}
\makeatletter
\patchcmd{\ps@pprintTitle}
  {Preprint submitted to}
  {}
  {}{}
\makeatother
\definecolor{fraunhofer_green}{HTML}{B4DCD3}
\definecolor{fraunhofer_gray}{HTML}{C7CACC}
\definecolor{fraunhofer_light_blue}{HTML}{D4E6F4}

\colorlet{punct}{red!60!black}
\definecolor{background}{HTML}{EEEEEE}
\definecolor{delim}{RGB}{20,105,176}
\colorlet{numb}{magenta!60!black}

\lstdefinelanguage{json}{
    basicstyle=\scriptsize\ttfamily,
    numbers=left,
    numberstyle=\scriptsize,
    stepnumber=1,
    numbersep=8pt,
    showstringspaces=false,
    breaklines=true,
    frame=lines,
    backgroundcolor=\color{background},
    literate=
     *{0}{{{\color{numb}0}}}{1}
      {1}{{{\color{numb}1}}}{1}
      {2}{{{\color{numb}2}}}{1}
      {3}{{{\color{numb}3}}}{1}
      {4}{{{\color{numb}4}}}{1}
      {5}{{{\color{numb}5}}}{1}
      {6}{{{\color{numb}6}}}{1}
      {7}{{{\color{numb}7}}}{1}
      {8}{{{\color{numb}8}}}{1}
      {9}{{{\color{numb}9}}}{1}
      {:}{{{\color{punct}{:}}}}{1}
      {,}{{{\color{punct}{,}}}}{1}
      {\{}{{{\color{delim}{\{}}}}{1}
      {\}}{{{\color{delim}{\}}}}}{1}
      {[}{{{\color{delim}{[}}}}{1}
      {]}{{{\color{delim}{]}}}}{1},
}

\title{Fed-DART and FACT: A solution for Federated Learning in a production environment}

\author[add1]{Nico Weber}
\author[add1]{Patrick Holzer}
\author[add1]{Tania Jacob}
\author[add1]{Enislay Ramentol}

\address[add1]{Fraunhofer Institute for Industrial Mathematics ITWM, Fraunhofer-Platz 1, 67663 Kaiserslautern, Germany}
\date{April 2022}

\begin{document}

\begin{abstract}
 Federated Learning as a decentralized artificial intelligence (AI) solution solves a variety of problems in industrial applications. It enables a continuously self-improving AI, which can be deployed everywhere at the edge. However, bringing AI to production for generating a real business impact is a challenging task. Especially in the case of Federated Learning, expertise and resources from multiple domains are required to realize its full potential. Having this in mind we have developed an innovative Federated Learning framework FACT based on Fed-DART, enabling an easy and scalable deployment, helping the user to fully leverage the potential of their private and decentralized data.
\end{abstract}

\maketitle

\section{Introduction}\label{sec:introduction}

\subsection{Federated Learning}
Over the last decade, the amount of data has grown almost exponentially \citep{statistaData}, and so far there is no end in sight to this momentum. This is due, among other things, to the steadily increasing number of technical devices as well as the no less increasing omnipresence of digital platforms and apps. Digital end devices collect data from almost all areas of our lives, machines, especially those in industry, are becoming increasingly networked and smarter, and digital platforms are becoming more and more important for both the economy and society. This has led to an era of machine learning (ML) and artificial intelligence (AI), for which a large availability of training data is essential to be successful in a productive environment.

However, as the volume of available data increased, so did the requirements for data protection and data security. The most notable of these is certainly the General Data Protection Regulation (GDPR) adopted by the European Union in 2018 \citep{gdpr}.
In many use cases, it is therefore not possible to copy the data stored on different instances and devices to a central server for training, yet there is often even a general interest in creating machine-learned models using the numerous data available. To resolve this conflict, McMahen et al. \citep{mcmahenblogfl, mcmahan2017communication} proposed an approach called \textit{federated learning} (FL) in 2017, in which the data remain local to the devices and yet a (global) model can be trained on them.

\begin{figure}[ht!]
\centering
\includegraphics[scale=1.0]{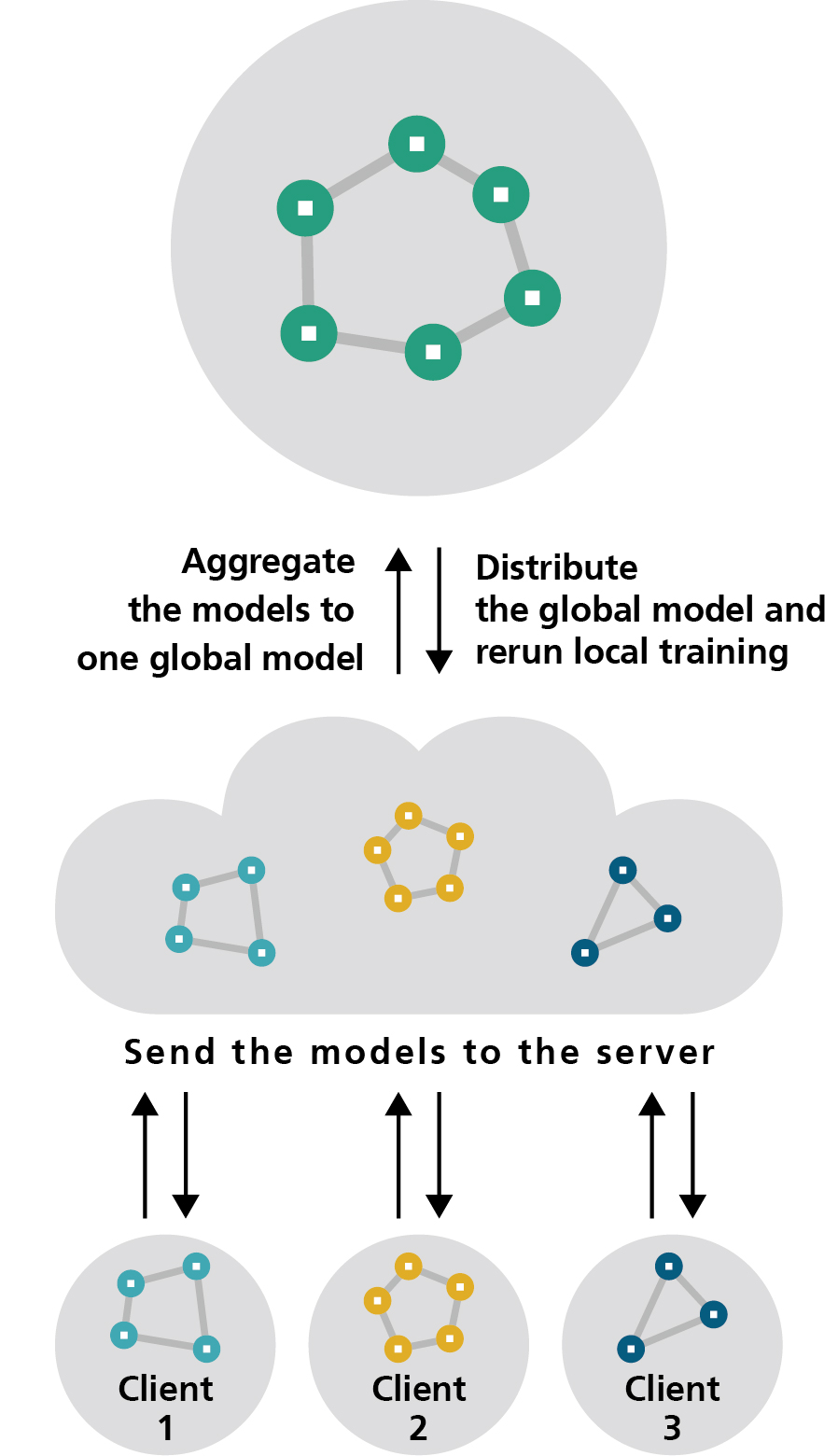}
\caption{Federated Learning scheme}
\label{fig:FL_schema}
\end{figure}

Roughly speaking, in the centralized version the idea is that there is a global ML model distributed to the single clients that is trained individually on the data-holding devices. These individually trained models, rather than the data itself, are then sent to the server and aggregated back into a global model. This process is then repeated for several rounds until the global model is sufficiently trained. The advantage is that the central server never sees the data itself, and that the data does not have to be independent and identically distributed. The process is sketched in Figure \ref{fig:FL_schema}.

For details see \citep{mcmahenblogfl, mcmahan2017communication}.

There are different types and settings of FL depending on the use case.
First of all, one has to distinguish between the number and the size of clients \citep{kairouz2021advances}.
\begin{itemize}
    \item 
    \textbf{Cross-device FL}: 
    In the cross-device setting there is typically a large number of relatively small clients, like mobile or IoT-devices.
    \item  
    \textbf{Cross-silo FL}: 
    In the cross-silo setting there is typically a small number of larger clients involved, usually around 2-100 clients. Clients in that case would be for example organizations or data centres.
\end{itemize}
Secondly, there is a distinction to be made according to the way in which the data is divided between the clients \citep{yang2019federated}.
\begin{itemize}
    \item 
    \textbf{Horizontal FL}: 
    In horizontal FL the clients hold data with the same features, but potentially different samples.
    \item  
    \textbf{Vertical FL}: 
    In vertical FL the clients hold data of the same samples, but potentially with different features.
    \item
    \textbf{Federated transfer learning}:
    Federated transfer learning is a mixture of both, i.e., the data of the clients can vary among the samples and the features.
\end{itemize}
In our work we focused on the setting of \textbf{centralized, horizontal, cross-silo FL} according to our intended use cases.

\subsection{Motivation}

The technical and algorithmic challenges of FL depend heavily on the specific requirements of each use case. In order to meet these requirements, it is of fundamental advantage to have the entire software stack under control.  Therefore, we have developed the central components in a holistic approach:  Our own scalable and fault-tolerant FL runtime (Fed-DART) combined with the required algorithms for enabling a sophisticated FL workflow (FACT). Besides having great flexibility and being framework agnostic, we have further points to emphasize:
\begin{itemize}
    \item	\textbf{Support for Personalized FL}:
Training models that have good performance, even in the setting of heterogeneous data and hardware resources is challenging. In order to create customized global models, a fine-grained mapping of which client delivered which results is needed. This information is provided by Fed-DART and processed appropriately by FACT.
\item	\textbf{Seamless transition from rapid, local prototyping to deployment in a production environment}:
The development of new suitable FL algorithms is most easily implemented locally on a single system. However, implementing this later in distributed systems brings new requirements. To keep this gap as small as possible, the distributed workflow can be simulated almost completely in the local test system.
\item	\textbf{Easy integration into production systems}:
The modular, loosely-coupled approach of our software stack allows an easy integration into preexisting software infrastructure or a flexible extension with non-FL components.  The required flexibility is achieved through a microservice architecture.
\end{itemize}

\FloatBarrier
\subsection{Section Overview}
The remaining sections of this paper are organized as follows. In Section \ref{sec:FEDDART} we present the software architecture of Fed-DART and FACT, including a discussion on the motivation for our design decisions.
This is followed by Section \ref{sec:sim_FL}, where we describe how a centralized ML system can be easily converted to a FL system using our framework. The focus here is on the minimum requirements to implement a working system. In Section \ref{sec:deployment} we present a container-based deployment strategy, detailing how such a system can efficiently be managed in various environments. This is followed by the conclusion in Section \ref{sec:conclusions}.

\FloatBarrier
\section{Software Architecture}\label{sec:FEDDART}
\subsection{Fed-DART}
FL as a collaborative ML paradigm can be implemented with two different communication schemes, either centralized or decentralized. In the decentralized scheme, all clients coordinate themselves in a peer-to-peer manner to improve the global model.
The centralized scheme instead involves a central server and can be implemented in two subforms: Server-centric or client-centric. In the server-centric approach the server has an active role and decides when a client must execute learning on his own data. In the the client-centric approach the clients themselves decide when to train and upload the results to the server. 
Comparing the server-centric approach with the commonly known MapReduce scheme, the server takes over the Reduce part by aggregating the local parameters. However, no explicit mapping takes place as the data is collected and kept in-situ. \\
The highly scalable parallelism is also used in the area of High Performance Computing for parallelizing data-intensive applications on multi-core clusters. GPI-Space \cite{GPI-SPACE}, developed at Fraunhofer ITWM, is a software solution for dealing with those computations. Written in C++, GPI-Space separates the coordination, which describes dependencies between tasks, from the computation on data. Using Petri nets as the workflow description language, GPI-Space can represent arbitrary dependency graphs between tasks. These tasks are then executed on the available hardware resources. GPI-Space scales efficiently, up to thousands of compute nodes, by using sophisticated workflow parallelization and scheduling strategies. The distributed runtime system of GPI-Space is fault-tolerant, which means in FL terminology, that a client can connect or disconnect at any time, without stopping the execution of the workflow.\\
The Distributed Analytics Runtime (DART) \cite{DART} is a Python API for GPI-Space, enabling geographically distributed applications with a MapReduce-like workflow. A capability could refer to a specific geographical location, which allows GPI-Space to schedule the task to that particular location. DART is framework-agnostic, which allows the execution of recent Python environments.\\
FL, in contrast, only partially follows the MapReduce-like workflow, and also has some further special requirements. FL clients are the owners of their data and are not available for the execution of a task at any desired time. FL is a data-centric paradigm, where the data distributions of the clients strongly influence the performance of the global model. 
Fed-DART is therefore an adaptation and further development of DART to meet the special requirements of a FL runtime in the domain of a server-centric FL scheme. Fed-DART itself does not natively provide FL algorithms, as these are dependent on the specific use case.
The following design goals were followed for the development of Fed-DART:
\begin{itemize}
    \item \textbf{Easy to use}: Fed-DART is based on the idea of separating the algorithmic level from the runtime level. The algorithm developer can fully concentrate on developing FL algorithms without the need for expertise in distributed computing. Experimental research can be done on a single-node with a seamless transition to multiple nodes in production systems.
    \item \textbf{Easy to integrate}: Fed-DART supports all Python-based ML frameworks. Converting a previously centralized learning workflow to  a FL one can be done very easily.
    \item \textbf{Easy to customize}: Fed-DART takes into account the flexible and diverse requirements of the specific use case. This is made possible by different levels of granularity with regard to the FL workflow. The implementation of personalized FL is easily possible by evaluating the supplied meta-information of the clients.
\end{itemize}
\subsubsection{Overview of the main components of Fed-DART}
\begin{figure}[h!]
    \centering
    \centering
    \includegraphics[width=0.75\textwidth, trim = 6cm 16cm 2cm 2cm]{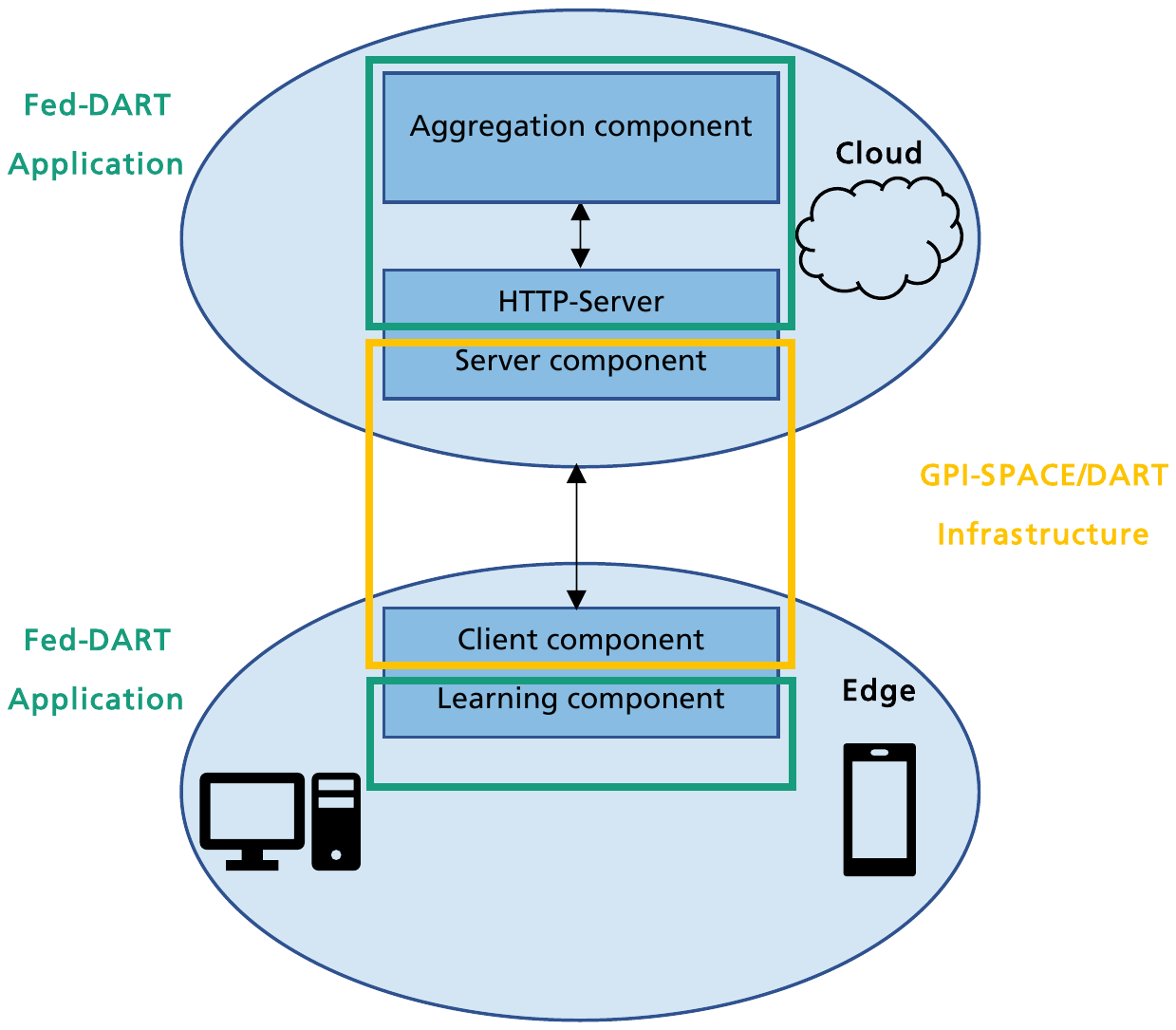}
    \caption{Three components of Fed-DART in a edge-device setting}
    \label{fig:three_comp}
\end{figure}
\begin{figure}[h!]
    \centering
    \centering
    \includegraphics[width=1\textwidth, trim = 1cm 15cm 2cm 2cm]{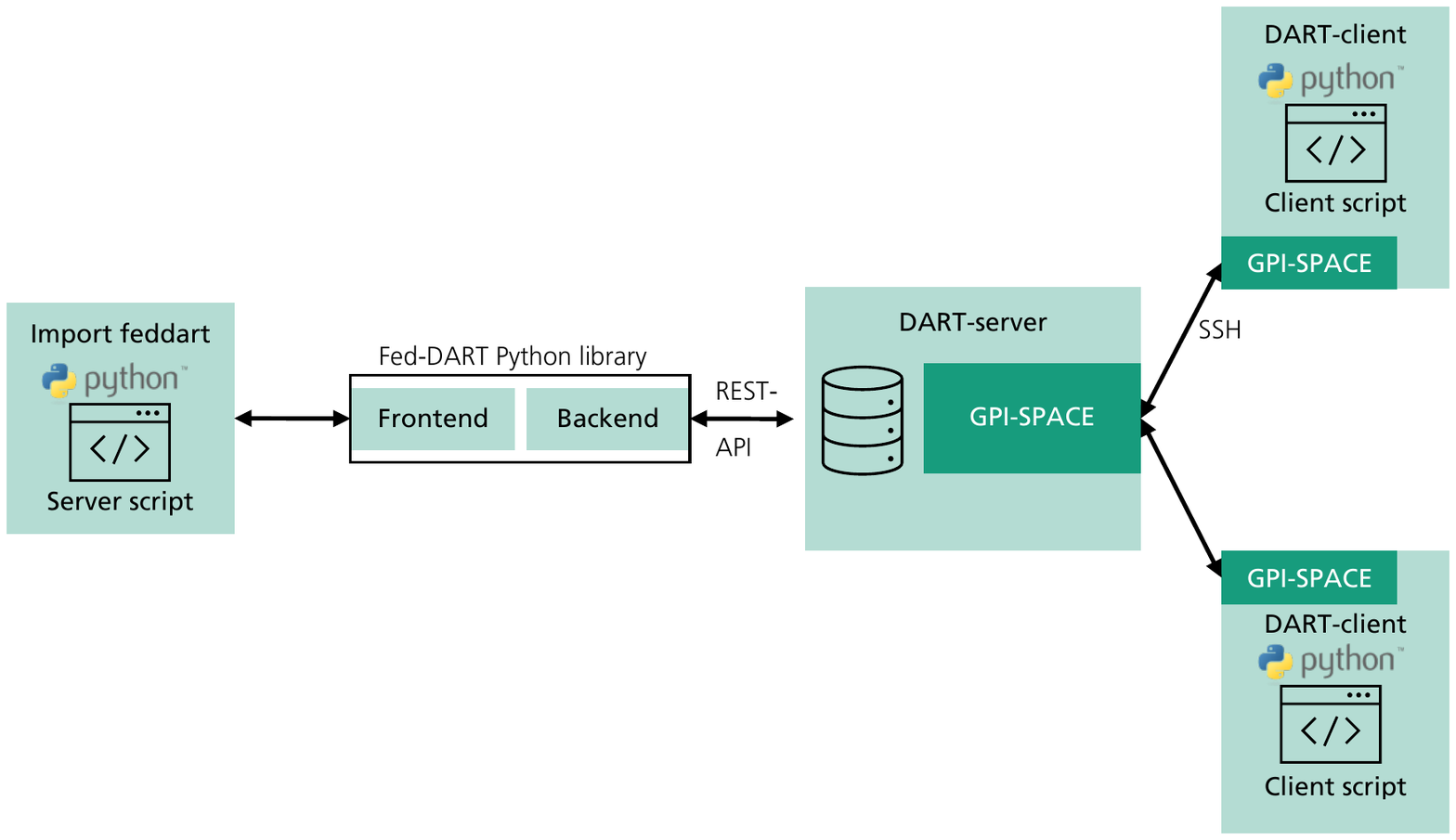}
    \caption{General workflow of Fed-Dart}
    \label{fig:fed_dart_workflow}
\end{figure}

Fed-DART consists of three separate components as shown in Figure \ref{fig:three_comp}, which enables separation of the FL application from the infrastructure. Typically the server and aggregation components run together in the cloud, whereas the client component is either deployed at the edge, such as on an embedded device, or on a larger system like a cloud or local compute instance. For a loose coupling between the DART backbone and the aggregation component, a https-server is introduced as an intermediate layer. 
\begin{itemize}
    \item \textbf{Server component}: The server component consists internally of two parts to achieve separation of concerns. A https-server handles the communication with the aggregation component over a REST-API. Furthermore, the https-server has an interface to manage the communication with DART. The server component of DART (\textbf{DART-Server}) orchestrates the clients and schedules the tasks to them.
    \item \textbf{Aggregation component}: This component is responsible for aggregating the parameters of the client models to one or multiple global models. The concrete implementation of the aggregation algorithm depends on the used ML framework and the statistical properties of the client data. Therefore it must be written by the user. The Fed-DART Python library, which runs in the aggregation component for communicating with the https-server, was developed to meet two main requirements. On the one hand the user needs a simple interface for interacting with the https-server, which abstracts the technical details away while allowing easy starting and analysis of the clients' learning routine. On the other hand the Fed-DART Python library must be scalable to handle the traffic of many clients and different tasks.

    \item \textbf{Client component}: The worker (\textbf{DART-client}) is responsible for executing the tasks and sending the results back to the DART-Server. The communication between DART-Server and DART-Client is SSH-secured. Provided that the server's public SSH-key is stored with a client, a client can connect to the server on its own during runtime or be added via IP address or hostname from the server component.
\end{itemize}
The general workflow for a FL use case is shown in Figure \ref{fig:fed_dart_workflow}.
The DART infrastructure together with the https-server must be set up once bare metal or as containers and can be reused for different FL use cases.
The aggregation component needs a use-case-specific Python script, where the Fed-DART Python library is imported.
In that script the FL workflow interacts with Fed-DART via the \texttt{WorkflowManager} for managing the tasks and the connected clients. A detailed description how the Fed-DART Python library can be used in the FL workflow is found in \ref{appendix_fed_dart_frontend}.
In the backend of the Fed-DART Python library the \texttt{Selector} is the central instance, which is responsible for orchestrating the communication with the DART-Server; details about the internal software design are given in \ref{appendix_fed_dart_backend}.
For simulating FL on a local system before implementing it as distributed system, the \texttt{test mode} of \texttt{WorkflowManager} can be activated.
In this mode a DART-Server together with DART-clients are simulated locally, which makes the algorithmic development and testing easier. This results in reduced development time and eases the adoption of FL.
When the code is deployed to a real client, again a use-case-specific script, whose functions the DART-client can call to execute a task, must be written. These functions should be annotated with \texttt{@feddart}. 
Moreover a configuration file for the DART-client is needed to specify the path of the Python environment, client script and the output directory for logging. 

\FloatBarrier
\subsection{FACT} 
\label{sec:FACT}
\defcitealias{tensorflow}{TensorFlow}
FACT (\textit{Federated Aggregation and Clustering Toolkit}) is a non public python library developed for the purpose of providing an easy-to-use toolkit for FL together with optional clustering algorithms. Similar publically available libraries are Flower, TensorFlow Federated and PySyft  \cite{beutel2020flower, tensorflow, Ziller2021PySyftAL}. However, these are still under construction and do not fit every purpose yet, especially when speaking about the usage in a productive environment and special tasks. The advantage of our self-developed FACT library, on the other hand, is that it is more lightweight and can be more easily adapted to specific use cases. Especially the dovetailing with the clustering is easier to maintain in an own library instead of changing internals of publically available libraries not built for that purpose. However, FACT was designed to provide general tools for FL, supporting multiple libraries such as Keras \cite{chollet2015keras}, Scikit-learn \cite{scikit-learn}, and others. It uses Fed-DART for communication between the server and the clients and the task handling.

\subsubsection{Software Design}
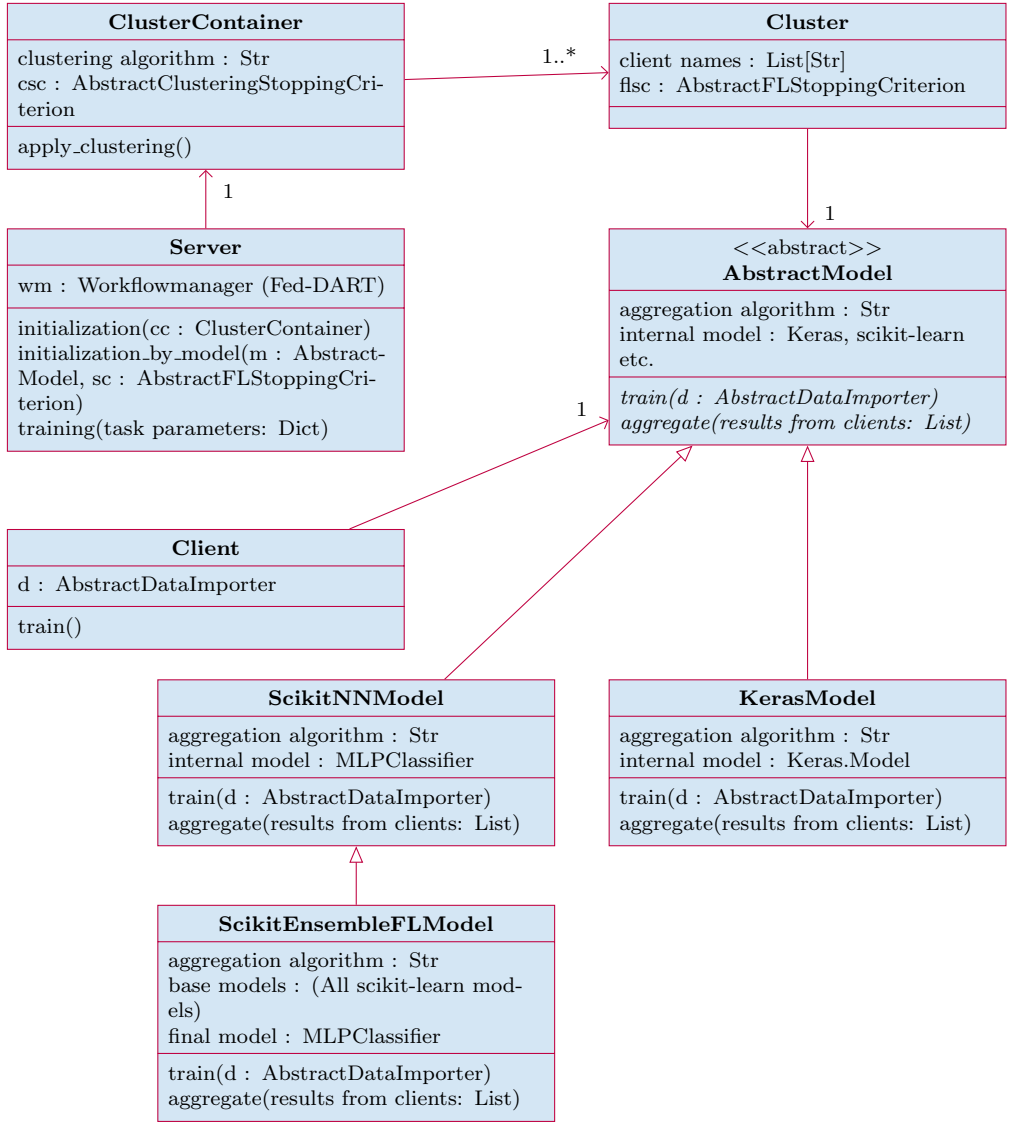
\begin{figure}
    \centering
    \begin{tikzpicture}[font=\scriptsize]
    
    \begin{class}[text width =5cm]{ClusterContainer}{0,0}
    \attribute{clustering algorithm : Str}
    \attribute{csc : AbstractClusteringStoppingCriterion}
    \operation{apply\_clustering()}
    \end{class}
    
    \begin{class}[text width =5cm]{Cluster}{8,0}
    \attribute{client names : List[Str]}
    \attribute{flsc : AbstractFLStoppingCriterion}
    \end{class}
    
    \begin{class}[text width =5cm]{Server}{0,-3}
    \attribute{wm : Workflowmanager (Fed-DART)}
    \operation{initialization(cc : ClusterContainer)}
    \operation{initialization\_by\_model(m : AbstractModel, sc : AbstractFLStoppingCriterion)}
    \operation{training(task parameters: Dict)}
    \end{class}
    
    \begin{class}[text width =5cm]{Client}{0,-7}
    \attribute{d : AbstractDataImporter}
    \operation{train()}
    \end{class}
    
    \begin{abstractclass}[text width =5cm]{AbstractModel}{8,-3}
    \attribute{aggregation algorithm : Str}
    \attribute{internal model : Keras, scikit-learn etc.}
    \operation[0]{train(d : AbstractDataImporter)}
    \operation[0]{aggregate(results from clients: List)}
    \end{abstractclass}
    
    \begin{class}[text width =5cm]{ScikitNNModel}{2,-9}
    \inherit{AbstractModel}
    \attribute{aggregation algorithm : Str}
    \attribute{internal model : MLPClassifier}
    \operation{train(d : AbstractDataImporter)}
    \operation{aggregate(results from clients: List)}
    \end{class}
    
    \begin{class}[text width =5cm]{KerasModel}{8,-9}
    \inherit{AbstractModel}
    \attribute{aggregation algorithm : Str}
    \attribute{internal model : Keras.Model}
    \operation{train(d : AbstractDataImporter)}
    \operation{aggregate(results from clients: List)}
    \end{class}
    
    \begin{class}[text width =5cm]{ScikitEnsembleFLModel}{2,-12}
    \inherit{ScikitNNModel}
    \attribute{aggregation algorithm : Str}
    \attribute{base models : (All scikit-learn models)}
    \attribute{final model : MLPClassifier}
    \operation{train(d : AbstractDataImporter)}
    \operation{aggregate(results from clients: List)}
    \end{class}
    
    \unidirectionalAssociation{ClusterContainer}{}{}{Cluster}{}{}
    \unidirectionalAssociation{Server}{}{}{ClusterContainer}{}{}
    \unidirectionalAssociation{Cluster}{}{}{AbstractModel}{}{}
    \unidirectionalAssociation{Client}{}{}{AbstractModel}{}{}
    
    \node at (4.7, -0.7) {1..*};
    \node at (8.3, -2.8) {1};
    \node at (0.3, -2.5) {1};
    \node at (5, -5.4) {1};

    \end{tikzpicture}
    \caption{Diagram of the mainly used classes in FACT}
    \label{fig:FACT_classes}
\end{figure}

A diagram of the mainly used classes together with their most important methods can be found in Figure \ref{fig:FACT_classes}.
The entry point for the user is the \texttt{Server} class. Internally it stores an instance of the \texttt{Workflowmanager} of Fed-DART to do the communication with the clients and sending tasks to them. The \texttt{Server} has two main methods, one for initializing the server and the clients and one to launch the training. 

To represent a physical client, there is a corresponding \texttt{Client} class. Each physical client needs to have a python file containing an initialization and a training method, which is called by Fed-DART and executes the corresponding code in the \texttt{Client} class. The \texttt{Client} class itself is responsible for the client-side code execution in FACT.

FACT can support different ML libraries such as Keras, Scikit-learn and others. This independence from the underlying library is achieved by introducing an abstraction layer with the \texttt{AbstractModel} class. The advantage is that it provides a consistent interface regardless of which library or model type is used. To support a new library or different types of models, one has to implement a class inheriting from \texttt{AbstractModel}. The aggregation algorithms, like federated averaging \citep{mcmahan2017communication} or FedProx \citep{li2020federated}, are part of the model class and the responsible \texttt{aggregate} method needs to be implemented for each model class, adapted to the internal model.

To support clustering in FACT, there are two further classes, the \texttt{ClusterContainer} and the \texttt{Cluster}. As the name indicates, the \texttt{ClusterContainer} is a container for the existing clusters, each represented by an instance of the \texttt{Cluster} class, responsible for orchestrating them. In particular, it is responsible for the clustering and when to stop. The instances of the  \texttt{Cluster} class, therefore, are responsible to store information about the contained clients and when to stop the FL on the cluster. Each cluster contains a central model, so instead of having one global model on the server there is one global model for each cluster.

The algorithmic and implementation details can be found in \ref{appendix:FACT}.

\FloatBarrier
\section{From Centralized Training to Simulated Federated Learning}\label{sec:sim_FL}

This section describes the process of building a federated version of an existing ML system using FACT and Fed-DART. The case of simulated FL (test mode) is described, where there are no actual distributed hardware devices, but the federated setting is simulated on a local test system. In Fed-DART, the test mode has the same workflow as the production mode so the conversion to a production system is then just a matter of configuration changes.

In a centralized ML task the procedure consists of the following main generic steps or components: 

\begin{enumerate}
    \item \textbf{Data loader}: A function or class to handle loading of the data. This includes any preprocessing and transformations of the data.
    \item \textbf{Model}: A class that defines a model architecture.
    \item \textbf{Training function}: A function that loops over batches of training data, calculates required metrics, and takes the required optimization steps.
    \item \textbf{Evaluation function}: A function that loops over the batches of test data and calculates required metrics.
\end{enumerate}

This centralized training scenario is the most common in ML systems and there are only minimal modifications necessary to convert existing centralized training code into a FL setup with FACT. In fact, the data loader, model and evaluation functions can be mostly reused as they are, with minimal code to integrate them into a FACT use case. Since the training function becomes more complicated in a federated setting it is mostly handled by FACT and only additional parameter setting and integration code needs to be implemented by the user. 

\begin{figure}[ht!]
\centering
    \includegraphics[width=1\textwidth, trim = 0cm 21cm 0cm 0cm]{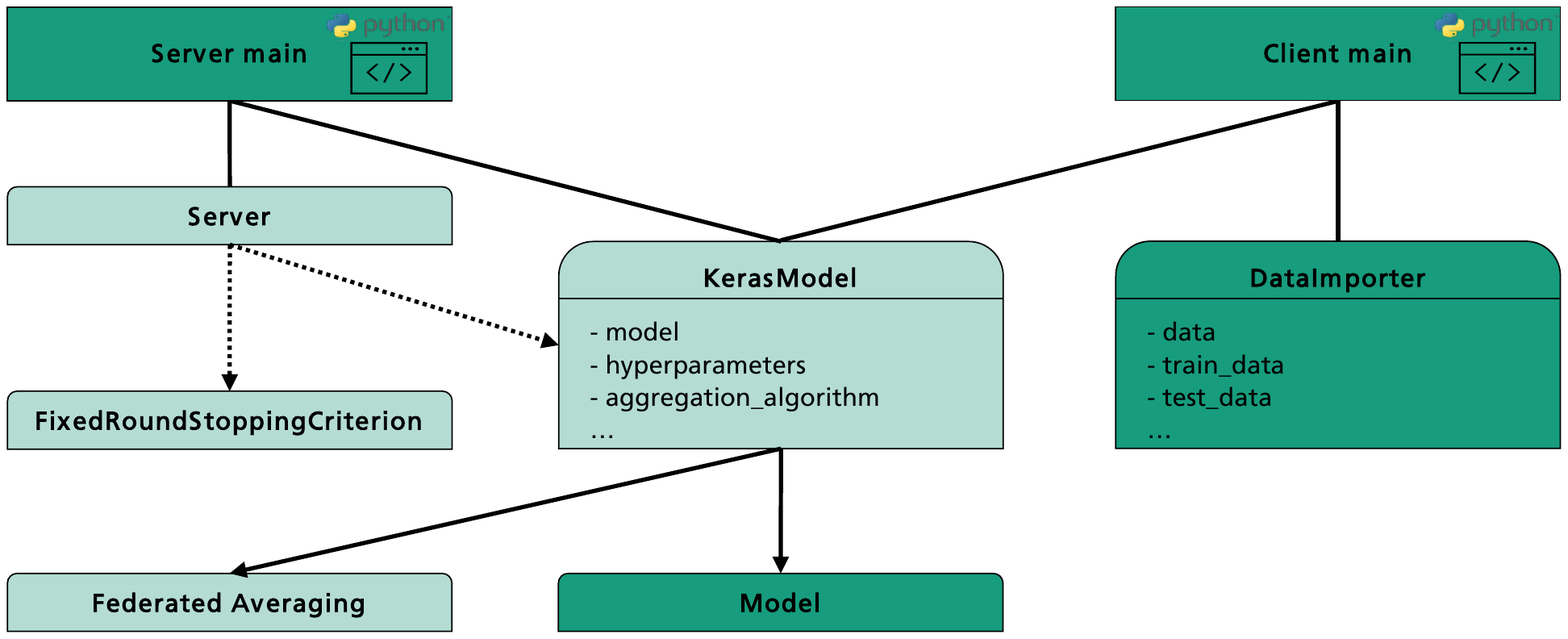}
\caption{Components that need to be implemented to convert from centralized to  with FACT are green. This example assumes the existing model is built with Keras.}
\label{fig:implement_simulated_fl}
\end{figure}

Figure \ref{fig:implement_simulated_fl} illustrates the essential components that need to be implemented for a new use case using FACT. Here the Keras framework is used as an example, however, this can be replaced with any other ML framework supported by FACT, or by a custom extension of FACT. Of the essential components, only the main server and client scripts need to be created from scratch. The main server script should, at minimum, instantiate a Server initialized with a FACT model and call the server's learning method. Other optional steps, such as exporting the trained model or performing some evaluation, can be included as required. For the main client script, a number of predefined functions for initialization, learning and (optionally) evaluation should be implemented. These will be called in order by FACT during training. The data importer and model can reuse much of the existing code from the classical ML system with minor refactoring to fit with FACT predefined methods. Further details about how these essential components can be implemented are available in \ref{appendix:simulatedFL}.

\section{Deployment as Cloud Native and Microservice Architecture}\label{sec:deployment}

Since Fed-DART and FACT were designed following a modular microservices architectual style, with each component performing specific functionality relatively independently of the rest, it lends itself nicely to being packaged and deployed using containers in a cloud-native manner. For this we support using the industry standard technologies, Docker \cite{docker} and Kubernetes \cite{kubernetes}, as described in the following subsections.
\subsection{Containerization}\label{subsec:containerization}
FACT provides a set of generic Docker images to support straightforward deployment and automation. There are separate Dockerfiles for server, client and aggregation images. All the images use Ubuntu, a Debian-based container image that provides a small base container image with the familiarity of a leading Linux distribution. Additionally, all images include Fed-DART, FACT and their dependencies. 

When instantiated, the server container runs a DART-server with various, user-definable ports opened to allow SSH and Fed-DART communications. Each client container starts a DART-client which connects to the server, also exposing the required ports. The aggregation container is used to initiate execution of a particular use case by calling the main server script. These three images provide a minimal base setup to support FL across distributed devices. The recommended way of adding additional dependencies and functionality to an image is to create a new Dockerfile using one of the provided generic images from FACT as a base.

\begin{figure}[h!]
    \centering
    \centering
    \includegraphics[width=1\textwidth, trim = 0cm 14cm 0cm 0cm]{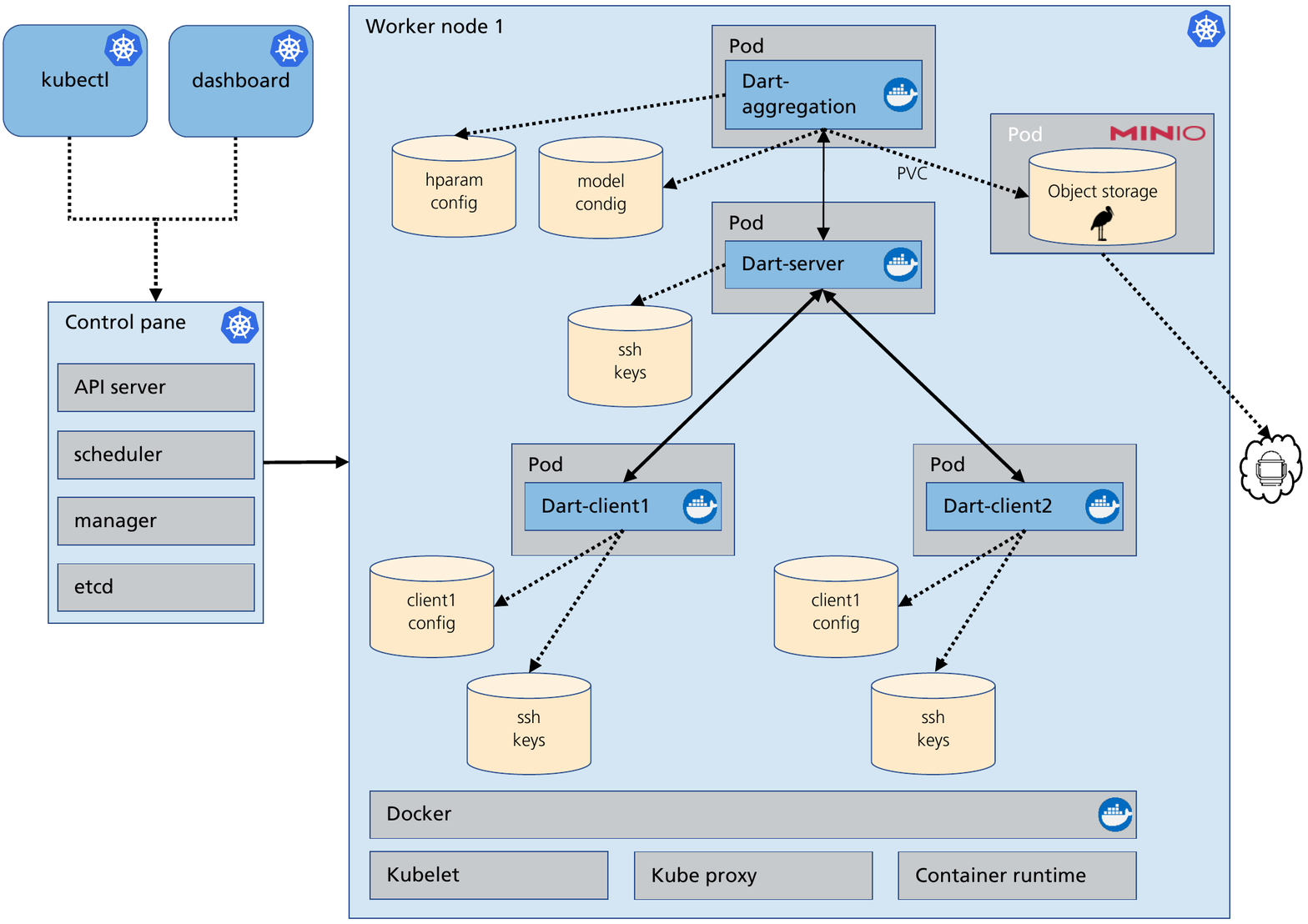}
    \caption{An example of a single-node Kubernetes cluster setup for a Fed-DART/FACT ML application.}
    \label{fig:microservices_arch}
\end{figure}

\subsection{Orchestration}
For container orchestration, applications using Fed-DART/FACT can easily take advantage of Kubernetes to have a single interface for deploying and managing containers in the cloud, on virtual machines or physical machines. Figure \ref{fig:microservices_arch} shows an example of a simple single-node Kubernetes cluster setup that could be used for development and testing purposes. This basic setup can be extended across multiple worker nodes and to include various other services as required. In addition to the Docker containers described in section \ref{subsec:containerization}, Figure \ref{fig:microservices_arch} also illustrates how MinIO, a distributed object storage server, can be integrated in order to, for example, save trained ML models to persistent S3 storage in the cloud. 

\subsection{CI/CD Pipeline}

\begin{figure}[h!]
    \centering
    \centering
    \includegraphics[width=0.75\textwidth, trim = 4cm 16cm 4cm 4cm]{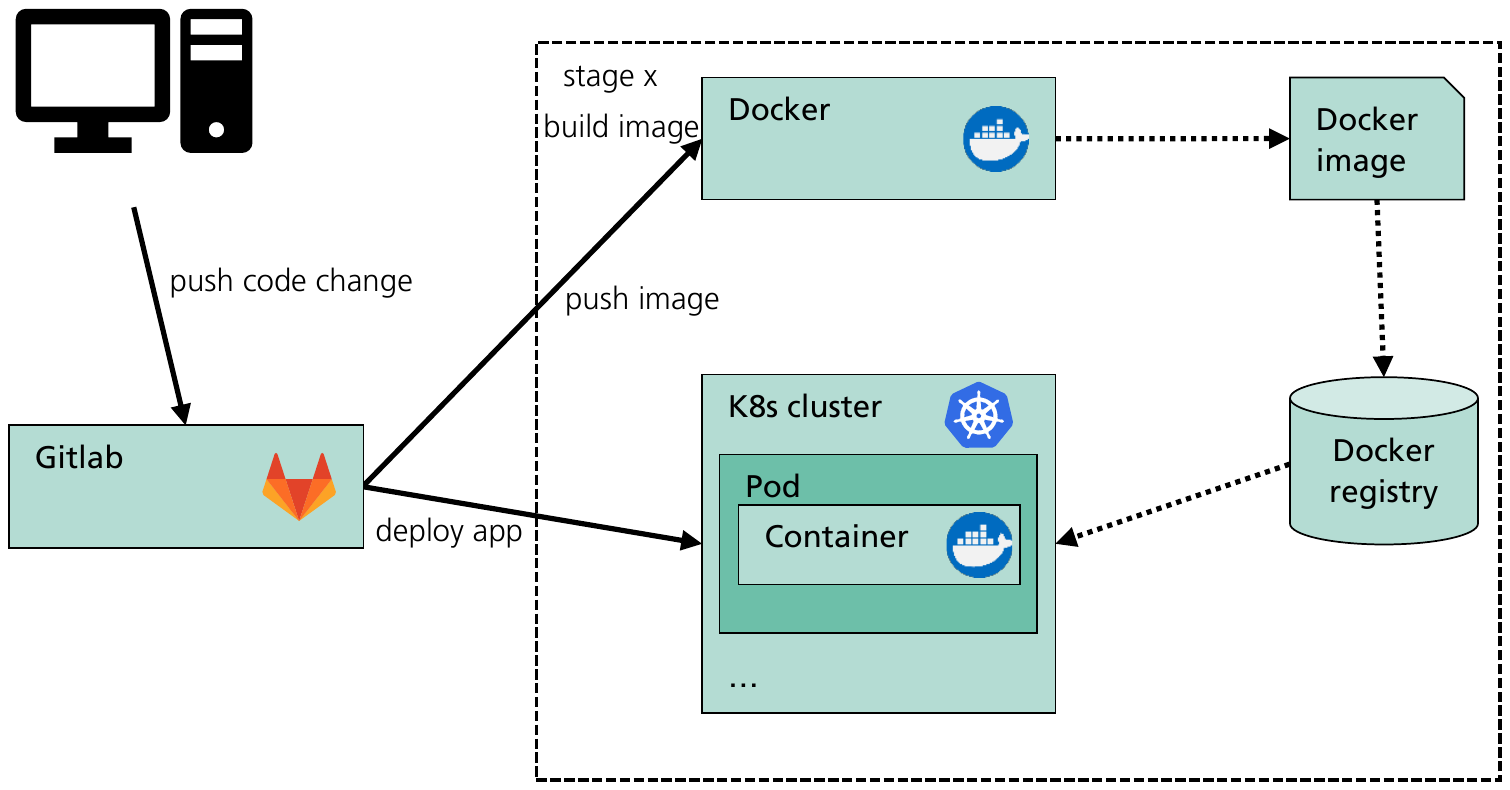}
    \caption{An illustration of one stage of a CI/CD pipeline, where a single microservice is updated, automatically built and deployed to a Kubernetes cluster.}
    \label{fig:cicd_pipeline}
\end{figure}

Having a microservices architecture and \emph{infrastructure as code} lends itself well to automation of the whole process, from development to deployment and maintenance, with a continuous integration/continuous delivery (CI/CD) pipeline. Figure \ref{fig:cicd_pipeline} illustrates one stage of the pipeline, where a a single microservice is updated, automatically built, pushed to a Docker registry and deployed to a Kubernetes cluster using Gitlab CI/CD. The process can be configured to automatically run unit tests, build different image versions for different environments, deploy to the various environments, and run integration tests in various stages from development and testing environments to production. There is typically one such block for each stage and the stages are progressed through sequentially, controlled by various triggers and checks, such as successful builds and passing of tests. There is typically one such pipeline for each of the microservices so they can be developed and deployed independently of each other.

\FloatBarrier

\section{ Conclusion}\label{sec:conclusions}
Each  use case is unique and has its own specific requirements, which can be only implemented flexibly enough with a holistic approach. The  runtime Fed-DART together with the FL toolkit FACT cover the entire development lifecycle for centralized, horizontal, cross silo FL  in a easy-to-use manner. The framework agnostic design enables great flexibility, using the most suited ML framework for the use case. At the beginning of each development lifecycle the focus is especially on developing suitable FL algorithms and workflows. FACT offers a variety of pre-implemented aggregation algorithms or if needed, new ones can be added easily through the modular design.  Fed-DART and FACT can simulate the distributed workflow completely in a local test system to enable rapid, local prototyping. With minor modifications it can be deployed in a real, distributed setting with the scalable and fault-tolerant Fed-DART runtime. Client-specific scheduling of tasks together with clustering algorithms enable the application of Personalized FL. The microservice architecture combined with prebuilt   Docker images allows easy integration in existing software architectures, helping the user to fully leverage the benefit of FL in business applications.  

\section*{Acknowledgements}\label{sec:acknowledgements}
This research was partly funded by the Bundesministerium für Wirtschaft und Energie (BMWi) grant number 01MK20013A.  
\newpage
\bibliographystyle{plain}
\bibliography{references}

\begin{thebibliography}{10}

\bibitem{docker}
Docker.
\newblock \url{https://www.docker.com}.

\bibitem{kubernetes}
Kubernetes.
\newblock \url{https://kubernetes.io}.

\bibitem{tensorflow}
Tensorflow federated.
\newblock \url{https://www.tensorflow.org/federated}.

\bibitem{beutel2020flower}
Daniel~J Beutel, Taner Topal, Akhil Mathur, Xinchi Qiu, Titouan Parcollet, and
  Nicholas~D Lane.
\newblock Flower: A friendly federated learning research framework.
\newblock {\em arXiv preprint arXiv:2007.14390}, 2020.

\bibitem{bonawitz2019towards}
Keith Bonawitz, Hubert Eichner, Wolfgang Grieskamp, Dzmitry Huba, Alex
  Ingerman, Vladimir Ivanov, Chloe Kiddon, Jakub Kone{\v{c}}n{\`y}, Stefano
  Mazzocchi, Brendan McMahan, et~al.
\newblock Towards federated learning at scale: System design.
\newblock {\em Proceedings of Machine Learning and Systems}, 1:374--388, 2019.

\bibitem{chollet2015keras}
Francois Chollet et~al.
\newblock Keras, 2015.

\bibitem{DART}
Fraunhofer ITWM.
\newblock Distributed analytics runtime ({DART}).
\newblock \url{https://github.com/cc-hpc-itwm/dart}, 2021.

\bibitem{Fed-DART}
Fraunhofer ITWM.
\newblock Distributed analytics runtime for federating learning.
\newblock \url{https://github.com/cc-hpc-itwm/fed-dart}, 2022.

\bibitem{kairouz2021advances}
Peter Kairouz, H~Brendan McMahan, Brendan Avent, Aur{\'e}lien Bellet, Mehdi
  Bennis, Arjun~Nitin Bhagoji, Kallista Bonawitz, Zachary Charles, Graham
  Cormode, Rachel Cummings, et~al.
\newblock Advances and open problems in federated learning.
\newblock {\em Foundations and Trends in Machine Learning}, 14(1--2):1--210,
  2021.

\bibitem{li2020federated}
Tian Li, Anit~Kumar Sahu, Manzil Zaheer, Maziar Sanjabi, Ameet Talwalkar, and
  Virginia Smith.
\newblock Federated optimization in heterogeneous networks.
\newblock {\em Proceedings of Machine Learning and Systems}, 2:429--450, 2020.

\bibitem{mcmahan2017communication}
Brendan McMahan, Eider Moore, Daniel Ramage, Seth Hampson, and Blaise~Aguera
  y~Arcas.
\newblock Communication-efficient learning of deep networks from decentralized
  data.
\newblock In {\em Artificial intelligence and statistics}, pages 1273--1282.
  PMLR, 2017.

\bibitem{mcmahenblogfl}
Brendan McMahan and Daniel Ramage.
\newblock Federated learning: Collaborative machine learning without
  centralized training data.
\newblock
  \url{https://ai.googleblog.com/2017/04/federated-learning-collaborative.html},
  Apr. 2017.

\bibitem{scikit-learn}
F.~Pedregosa, G.~Varoquaux, A.~Gramfort, V.~Michel, B.~Thirion, O.~Grisel,
  M.~Blondel, P.~Prettenhofer, R.~Weiss, V.~Dubourg, J.~Vanderplas, A.~Passos,
  D.~Cournapeau, M.~Brucher, M.~Perrot, and E.~Duchesnay.
\newblock Scikit-learn: Machine learning in {P}ython.
\newblock {\em Journal of Machine Learning Research}, 12:2825--2830, 2011.

\bibitem{GPI-SPACE}
Tiberiu Rotaru, Mirko Rahn, and Franz{-}Josef Pfreundt.
\newblock Mapreduce in gpi-space.
\newblock In Dieter an~Mey, Michael Alexander, Paolo Bientinesi, Mario
  Cannataro, Carsten Clauss, Alexandru Costan, Gabor Kecskemeti, Christine
  Morin, Laura Ricci, Julio Sahuquillo, Martin Schulz, Vittorio Scarano,
  Stephen~L. Scott, and Josef Weidendorfer, editors, {\em Euro-Par 2013:
  Parallel Processing Workshops - BigDataCloud, DIHC, FedICI, HeteroPar, HiBB,
  LSDVE, MHPC, OMHI, PADABS, PROPER, Resilience, ROME, and {UCHPC} 2013,
  Aachen, Germany, August 26-27, 2013. Revised Selected Papers}, volume 8374 of
  {\em Lecture Notes in Computer Science}, pages 43--52. Springer, 2013.

\bibitem{statistaData}
Statista.
\newblock Total data volume worldwide 2010-2025.
\newblock
  \url{https://www.statista.com/statistics/871513/worldwide-data-created/},
  March 2022.

\bibitem{gdpr}
European Union.
\newblock General data protection regulation.
\newblock \url{https://eur-lex.europa.eu/eli/reg/2016/679/oj}, 2018.

\bibitem{yang2019federated}
Qiang Yang, Yang Liu, Tianjian Chen, and Yongxin Tong.
\newblock Federated machine learning: Concept and applications.
\newblock {\em ACM Transactions on Intelligent Systems and Technology (TIST)},
  10(2):1--19, 2019.

\bibitem{Ziller2021PySyftAL}
Alexander Ziller, Andrew Trask, Antonio Lopardo, Benjamin Szymkow, Bobby
  Wagner, Emma Bluemke, Jean-Mickael Nounahon, Jonathan Passerat-Palmbach,
  Kritika Prakash, Nick Rose, Theo Ryffel, Zarreen~Naowal Reza, and G.~Kaissis.
\newblock Pysyft: A library for easy federated learning.
\newblock 2021.

\end{thebibliography}
\newpage

\newpage
\appendix
\section{Fed-DART}
The following appendix section provides a detailed description of the Fed-DART Python library from the user perspective and the internal software design of the library.
The Fed-DART Python library is available as a local Python package \cite{Fed-DART} and can be installed via pip.
\subsection{Interaction of the user with the Fed-DART Python library}\label{appendix_fed_dart_frontend}
To create a script on the aggregation side  and to interact with Fed-DART, the first step is to instantiate a \texttt{WorkflowManager}.
The workflow of Fed-DART can be segmented into two phases: the starting phase and the learning phase.
The Fed-DART workflow in the starting phase is shown in Alg. \ref{alg:init_fed_dart}.
The user can optionally submit in the starting phase, an \texttt{initTask} to the \texttt{workflowManager} before the actual FL begins. Fed-DART guarantees, that this initialization function is executed on each client before other tasks can run.
\begin{algorithm}[h!]
\caption{Starting phase in Fed-DART}
\label{alg:init_fed_dart}
\SetKwInOut{Input}{Input}\SetKwInOut{Output}{Output}
\Input{\textbf{Server file} and optional \textbf{client file}}
Initialize the \textbf{workflow manager}\;
\If{clients must be initialized}{
        \textbf{Workflow manager} creates an \textbf{init task} for setting up the clients. Typically the  model structure is passed via the \textbf{parameterDict}.
    }
Use the \textbf{workflowManager} and the \textbf{server file} to connect to the DART-Server. The \textbf{init task} is scheduled to all clients, the optional \textbf{client file} is used to bootstrap further \textbf{DART-Clients}. \
Wait until initialization phase is finished. 
\end{algorithm}
Every task type in \texttt{workflowManager} has a similar interface with at least three arguments as seen in Listing \ref{code:task}:
\begin{enumerate}
    \item \texttt{parameterDict}: In case of the default task \texttt{parameterDict} contains all client names as keys, where the task should be executed. The associated value is again a dictionary, containing the function arguments of \texttt{executeFunction}.
    \item \texttt{filePath}: Path to the client script. 
    \item \texttt{executeFunction}: Name of the function that should be executed by the task. The \texttt{executeFunction} must be located in \texttt{filePath} and annotated with \texttt{@feddart}.
\end{enumerate}
\begin{lstlisting}[frame = single, language=python, caption= Code snippet with init and default task, label=code:task]
from feddart.workflowManager import WorkflowManager

wm = WorkflowManager()
pD = {"model_structure": global_model_structure}
wm.createInitTask( parameterDict = pD
                 , filePath = "client_script"
                 , executeFunction = "init"
                 )
....
pD = { "client1": { "weights": global_weights
                  , "epochs": 10
                  }
     }
handle = wm.startTask( parameterDict =  pD
                     , filePath = "client_script"
                     , executeFunction = "learn"
                     )
# handle is non-blocking
# continue with further computations
....
\end{lstlisting}

\begin{algorithm}[h!]
\caption{Learning phase in Fed-DART}
\label{alg:learn_fed_dart}
\SetKwInOut{Input}{Input}\SetKwInOut{Output}{Output}
\Input{\textbf{Workflow manager}} 
\ForEach{\textbf{learning round}}{
    Get the connected clients from \textbf{workflow manager}\;
    Define the learning parameters for each client and store them in the \textbf{parameterDict}\;
    Pass \textbf{parameterDict} to \textbf{workflow manager} and start a task\;
    \If{Start task was successfull}
    {
    Return \textbf{handle} as a unique identifier for the task\;
    }
    \Else{Throw an error that the task was not valid\;}
    \While{task not finished or maximal waiting time not reached}
    {
    Use \textbf{handle} to get current task status from \textbf{workflow manager}\;
    }
    Get available \textbf{taskResults} with \textbf{handle} from \textbf{workflow manager}. Analyse the results and perform a new learning round if necessary\;
}
\end{algorithm}

After establishing a connection to the DART-Server the user can query the connected clients and continue with the learning phase, as shown in Alg. \ref{alg:learn_fed_dart}.
Based on that information a learning task with client-specific parameters can be submitted to the DART-server. If the task was accepted, a handle is returned to the user. Since Fed-DART is non-blocking, this handle allows the user to continue with their workflow and query the status of the task as required.
Finally, at the user's convenience, the current available task results can be downloaded. This means, in particular, that there is no need to wait until all participating clients have finished executing the task. The task results are returned as a list, with each list element having the following attributes
\begin{itemize}
    \item \texttt{taskResult.deviceName}: The name of the client
    \item \texttt{taskResult.duration}: Information on how long the client needed to process the task (in seconds)
    \item \texttt{taskResult.resultDict}: The result of the client in dictionary format, for example \texttt{\{"result\_0":5, "result\_1":2\}} 
    \item \texttt{taskResult.resultList}: The result of the client in list format, for example \texttt{[5,2]}
\end{itemize}
The meta-information \texttt{deviceName} and \texttt{duration} can be used in further workflow steps for fine-granular FL such as clustering of similar clients.

\subsection{Software design of the Fed-DART Python library} \label{appendix_fed_dart_backend}
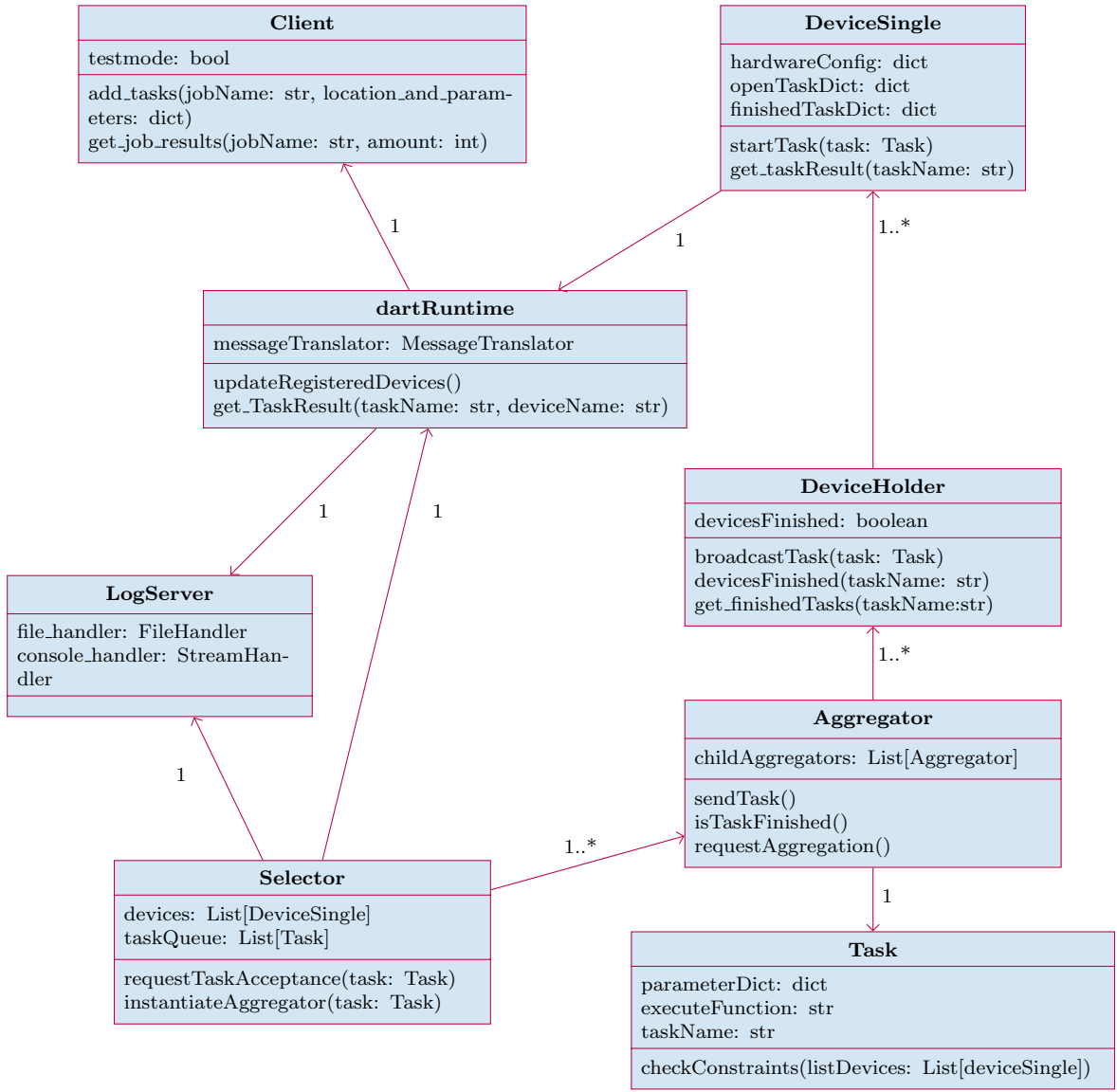
\begin{figure}
    \centering
    \begin{tikzpicture}[font=\scriptsize]
    
    \begin{class}[text width =5cm]{Selector}{0,-12}
    \attribute{devices: List[DeviceSingle]}
    \attribute{taskQueue: List[Task]}
    \operation{requestTaskAcceptance(task: Task)}
    \operation{instantiateAggregator(task: Task)}
    \end{class}
    
    \begin{class}[text width =4cm]{LogServer}{-2,-8}
    \attribute{file\_handler: FileHandler}
    \attribute{console\_handler: StreamHandler }
    \end{class}
    
    \begin{class}[text width =5cm]{Aggregator}{8,-9.75}
    \attribute{childAggregators: List[Aggregator]}
    \operation{sendTask()}
    \operation{isTaskFinished()}
    \operation{requestAggregation()}
    \end{class}
    
    \begin{class}[text width =5cm]{DeviceHolder}{8,-6.5}
    \attribute{devicesFinished: boolean}
    \operation{broadcastTask(task: Task)}
    \operation{devicesFinished(taskName: str)}
    \operation{get\_finishedTasks(taskName:str)}
    \end{class}
    
    \begin{class}[text width =6.5cm]{Task}{8,-13}
    \attribute{parameterDict: dict}
    \attribute{executeFunction: str}
    \attribute{taskName: str}
    \operation{checkConstraints(listDevices: List[deviceSingle])}
    \end{class}
    
    \begin{class}[text width =4cm]{DeviceSingle}{8,0}
    \attribute{hardwareConfig: dict}
    \attribute{openTaskDict: dict}
    \attribute{finishedTaskDict: dict}
    \operation{startTask(task: Task)}
    \operation{get\_taskResult(taskName: str)}
    \end{class}
    
    \begin{class}[text width =6.5cm]{dartRuntime}{2,-4}
    \attribute{messageTranslator: MessageTranslator }
    \operation{updateRegisteredDevices()}
    \operation{get\_TaskResult(taskName: str, deviceName: str)}
    \end{class}
    
    \begin{class}[text width =6cm]{Client}{0,0}
    \attribute{testmode: bool}
    \operation{add\_tasks(jobName: str, location\_and\_parameters: dict)}
    \operation{get\_job\_results(jobName: str, amount: int)}
    \end{class}

    \unidirectionalAssociation{Selector}{}{}{Aggregator}{}{}
    \unidirectionalAssociation{Aggregator}{}{}{Task}{}{}
    \unidirectionalAssociation{Aggregator}{}{}{DeviceHolder}{}{}
    \unidirectionalAssociation{DeviceHolder}{}{}{DeviceSingle}{}{}
    \unidirectionalAssociation{DeviceSingle}{}{}{dartRuntime}{}{}
    \unidirectionalAssociation{Selector}{}{}{dartRuntime}{}{}
    \unidirectionalAssociation{Selector}{}{}{LogServer}{}{}
    \unidirectionalAssociation{dartRuntime}{}{}{LogServer}{}{}
    \unidirectionalAssociation{dartRuntime}{}{}{Client}{}{}
    
    \node at (3.9, -11.8) {1..*};
    \node at (8.2, -12.5) {1};
    \node at (8.3, -9.1) {1..*};
    \node at (8.3, -3.1) {1..*};
    \node at (5.3, -3.3) {1};
    \node at (1.3, -3.1) {1};
    \node at (0.3, -7.1) {1};
    \node at (1.9, -7.1) {1};
    \node at (-1.7, -10.8) {1};
    \end{tikzpicture}
    \caption{Diagram of the main classes in Fed-DART}
    \label{fig:Fed_DART_classes}
\end{figure}

\begin{figure}[h!]
    \centering
    \centering
    \includegraphics[width=1\textwidth, trim = 5cm 15cm 2cm 2cm]{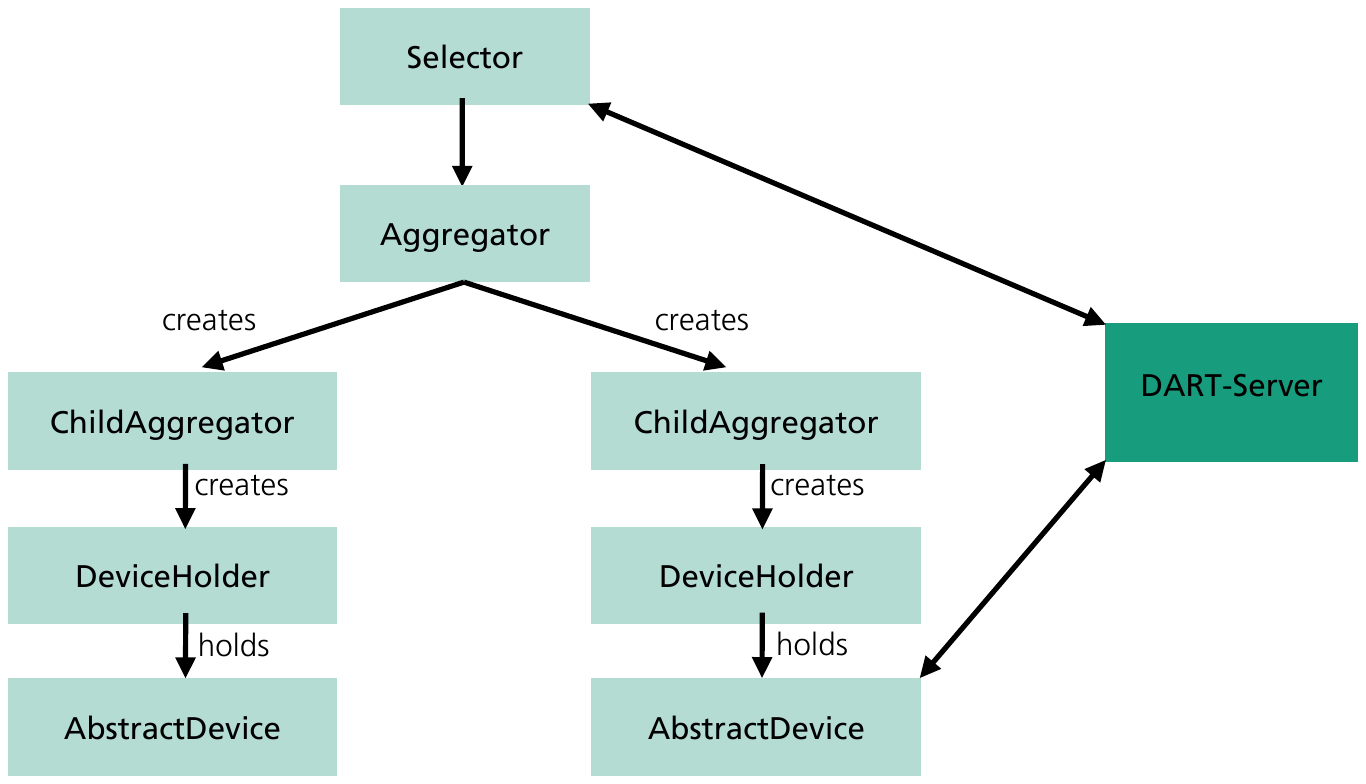}
    \caption{Creation of ephemeral classes after accepting a task from the selector}
    \label{fig:creation_ephermal_classes}
\end{figure}

In the Fed-DART software design, the most important classes are \texttt{WorkflowManager} and \texttt{Selector}. As explained in the previous subsection, the \texttt{WorkflowManager} is the central instance with which the end user interacts.
Figure \ref{fig:Fed_DART_WorkflowManager} lists the most important functions of the \texttt{WorkflowManager}.
\begin{figure}[h!]
    \centering
    \begin{tikzpicture}[ font=\scriptsize]
    
    \begin{class}[text width =5cm]{WorkflowManager}{4,0}
    \attribute{selector: Selector }
    \attribute{initTask: InitTask}
    \attribute{testMode: bool}
    \attribute{currentDeviceNames: List[str]}
    \attribute{logger: LogServer}
    \operation{createInitTask()}
    \operation{startFedDART()}
    \operation{getAllDeviceNames()}
    \operation{startTask()}
    \operation{getTaskStatus()}
    \operation{getTaskResult()}
    \operation{stopTask()}
    \end{class}
    \end{tikzpicture}
    \caption{Main attributes of the class \texttt{WorkflowManager} and most import functions for the user.}
    \label{fig:Fed_DART_WorkflowManager}
\end{figure}
The \texttt{WorkflowManager} communicates with the \texttt{Selector}, which is the central element in the internal software design of the Fed-DART Python library, enabling easy and scalable orchestration of the clients.
The internal Fed-DART system design is inspired by \cite{bonawitz2019towards}.
An overview of the internal software design is given in Figure \ref{fig:Fed_DART_classes}.
The classes can be grouped into two categories: ephemeral and non-ephemeral.\\
\\
{\large\textbf{Non-ephemeral classes}}\\
\\
\textbf{Selector} has knowledge about the connected clients and is responsible for accepting or rejecting incoming task requests from the \texttt{WorkflowManager}. It schedules the \texttt{initTask} to new clients. If a task request is accepted, the task is put into a queue until the DART-Server has capacity to schedule a new task. After scheduling a task, the \texttt{WorkflowManager} creates an \texttt{Aggregator} and hands over the \texttt{DevciceSingle}s to them. It manages all existing \texttt{Aggregators}. \\
\\
\textbf{DartRuntime} has the function of a helper class to translate \texttt{DeviceSingle}'s requests into a compliant format for the REST client. In the other direction, the incoming traffic from the REST client is decoded.\\
\\
\textbf{Client} communicates with the DART-Server via the REST-API.
If the test mode of the client is active, a dummy DART-Server is simulated, which handles the REST-API communication. In this case the dummy DART-Server executes the task in a sequential manner on the local machine. \\
\\
\textbf{LogServer} logs the communication between the DART-Server and the involved classes on the Fed-DART side. The user can specify different log levels. Especially for debugging distributed systems it is of essential advantage to have this information. \\
\\
\textbf{DeviceSingle} is the virtual representation of each real physical client. It handles its attributes including IP address, hostname, and hardware configuration. All communication to the DART-server related to a specific physical client goes through an instance of \texttt{DeviceSingle}. Each \texttt{deviceSingle} caches the task parameters of an open task and the task results of already finished tasks.\\

\ \\
{\large\textbf{Ephemeral classes}}
\ \\
\ \\
The creation cycle of the ephemeral classes after a task has been accepted by the \texttt{selector} is shown in Figure \ref{fig:creation_ephermal_classes}. \\
\\
\textbf{Aggregator} is responsible for managing a task. In order to scale with the amount of clients required for a task, the Aggregator can spawn \texttt{ChildAggregator}s to create a tree structure. This allows balancing and  parallelization of operations if needed. The associated clients are stored in one or more \texttt{deviceHolders}.
The \texttt{Aggregator} is the central instance to query or manipulate the task status. \\
\\
\textbf{DeviceHolder} groups multiple \texttt{DevinceSingle}s together.
Every request to a client must go through the \texttt{DeviceHolder}. 
If possible, computations or requests are performed on \texttt{deviceHolder} level to avoid too many small operations on \texttt{deviceSingle} level.\\
\\
\textbf{Task} manages all relevant information, such as the function to be executed and the function parameters for each client. A check function verifies the task requirements to ensure that hardware requirements and device availability are fulfilled.

\section{FACT}
\label{appendix:FACT}
The following appendix section provides a detailed description of the FACT Python library.
\subsection{Initialization}
The initialization method of the \texttt{Server} is described in algorithm \ref{alg:FACT_INIT}.
The purpose of it is to set all necessary parameters like the model, the clusters, the used clustering and aggregation algorithms, the stopping criteria etc. In fact, there are two different initialization methods implemented, which are selected depending on whether one would like to apply standard FL without clustering or to use the clustering capabilities. If initialized without clustering, i.e., if only a model is given, internally a \texttt{ClusterContainer} with one single \texttt{Cluster} holding that model is created. Furthermore, the stopping criterion is set to stop after one round of clustering and the clustering algorithm is set to do nothing. In that case, the setup is equivalent to standard FL.
\\
\\
\begin{algorithm}[H]
\caption{Initialize routine in FACT}
\label{alg:FACT_INIT}
\SetKwInOut{Input}{Input}\SetKwInOut{Output}{Output}
\Input{A \textbf{cluster container} or a \textbf{model} I.}
\If{I is a \textbf{model}}{
        $I\gets$ Create a new \textbf{cluster container} containing a single \textbf{cluster} which  all \textbf{clients} are part of. The \textbf{clustering algorithm} is set to static, the \textbf{clustering stopping criterion} is set to one fixed round and the \textbf{model} of the single \textbf{cluster} is given by I\;
    }
Start Fed-DART\;
\ForEach{\textbf{cluster} in the \textbf{cluster container}}{
    Initialize the local \textbf{models} on the \textbf{clients} corresponding to the \textbf{cluster} based on the global \textbf{model} in the \textbf{cluster}\;
}
\end{algorithm}

\FloatBarrier
\subsection{Training}
\FloatBarrier
The outer procedure of the training method of the server is described in algorithm \ref{alg:FACT_LEARN}, which handles the clustering in FACT. Roughly speaking, it executes a training session on each cluster, described in algorithm \ref{alg:FACT_TRAIN_CLUSTER}, which is equivalent to standard FL with only the clients corresponding to that cluster. After that it applies the clustering algorithm and proceeds from the beginning, if the clustering stopping criterion is not satisfied. So the clustering adds only a further loop to iterate over, to standard FL.

\begin{algorithm}[H]
\caption{Learning routine in FACT}
\label{alg:FACT_LEARN}
\SetKwInOut{Input}{Input}\SetKwInOut{Output}{Output}
\SetKwProg{ForAllInParallel}{foreach}{ do in parallel}{end} 
\Input{Task parameters.}
\ForEach{clustering round in $\mathbb{N}$}{
    \ForAllInParallel{\textbf{cluster} in the \textbf{cluster container}}{
        train\_cluster(\textbf{cluster}, task parameters, clustering round)\;
    }
    Apply the clustering algorithm on the cluster container\;
    \If{\textbf{clustering stopping criterion} is satisfied}{
        stop\;
    }
}
\end{algorithm}

\begin{algorithm}[H]
\caption{Training a cluster in FACT}
\label{alg:FACT_TRAIN_CLUSTER}
\SetKwInOut{Input}{Input}\SetKwInOut{Output}{Output}
\SetKwProg{ForAllInParallel}{foreach}{ do in parallel}{end} 
\Input{A \textbf{cluster} C, task parameters T and the current clustering round i.}
\ForEach{training round in $\mathbb{N}$}{
    \ForAllInParallel{\textbf{client} in the \textbf{cluster C}}{
        The \textbf{server} uses Fed-DART to send a training task to each \textbf{client} in the given \textbf{cluster} which triggers the train method with given parameters T of the \textbf{model} in the \textbf{client}\;
    }
    Fetch the updated model parameters from the \textbf{clients} after training and aggregate them on the \textbf{server} with the \textbf{cluster} side given method\;
    \If{\textbf{training stopping criterion} is satisfied}{
        stop\;
    }
}
\end{algorithm}

\FloatBarrier
\subsection{Implemented Model Classes}
\FloatBarrier

Currently there are three concrete implementations of the \texttt{AbstractModel} class.
\begin{itemize}
    \item 
    \textbf{KerasModel}: Supports Keras models. Implemented aggregation algorithms: (Weighted) federated averaging and FedProx.
    \item
    \textbf{ScikitNNModel}: Supports the usage of the MLPClassifier from Scikit-learn. Implemented aggregation algorithm: (Weighted) federated averaging.
    \item
    \textbf{ScikitEnsembleFLModel}: We introduced a new method named \textbf{ensemble FL} to use further model types for FL which makes use of the stacking technique. It allows to use arbitrary ML models like decision trees, random forests, support vector machine etc. in a federated setup. The details will be discussed in a further paper.
    
    Implemented aggregation algorithm: It inherits the aggregation algorithms from \textbf{ScikitNNModel} via applying the aggregation only to the final model.
\end{itemize}

To support further libraries and models, one only has to implement a subclass of \texttt{AbstractModel} and all of its abstract methods.

\FloatBarrier
\subsection{Implemented Stopping Criteria}
\FloatBarrier

There are two different types of stopping criteria, one for the clustering and one for the FL. 
For both there exists abstract base classes called
\begin{itemize}
    \item 
    \texttt{AbstractClusteringStoppingCriterion} and
    \item
    \texttt{AbstractFLStoppingCriterion}.
\end{itemize}
Currently we have only implemented one subclass of each, which check if the number of iteration exceeds a given value, i.e., to have a fixed number of iterations. 

To create new stopping criteria, one only has to implement a subclass of the abstract classes and all of its abstract methods. However, the server, which is responsible for the iterations in the training, passes only the current round number to these stopping criteria. If they need further information, such as how much the weights of the neural network have changed, this argument has to be added in the server code. Since the arguments are passed to the stopping criteria via keyword arguments, this would not affect the other existing implementations.

\section{From Centralized Training to simulated Federated Learning}
\label{appendix:simulatedFL}
\subsection{Server-side implementation}
\subsubsection{FACT Model} 
The FACT model can be an existing implementation such as FACT's \texttt{KerasModel} or a custom subclass of the \texttt{AbstractModel}. The FACT model classes accept instances of the respective framework's model class directly. For instance, the \texttt{KerasModel} expects to be passed an object of type \texttt{tf.keras. Model} upon initialization. The model itself can be hardcoded or loaded from configuration files of various supported formats, including JSON and YAML. Hyperparameters, such as the optimizer, loss, batch size, and number of local training epochs, can also be set in the FACT model. 

In addition to the standard hyperparameters of centralized training, a federated aggregation algorithm must be specified here. For this, the FACT models come with a number of built-in aggregation possibilities, including standard federated averaging.
\subsubsection{Server main script}
To create the main server script, the first step is to instantiate a  \texttt{Server}. The Fed-DART server and device configuration file paths must be specified here. The server configuration file must, at minimum, contain a key-value pair called \emph{server}, specifying the server address. This could be, for example, https://127.0.0.1:7777 in the test mode. A minimal example of a server device configuration file is given in Listing \ref{code:server_config}. 

\begin{lstlisting}[language=json, caption=Minimal example of a server configuration file., label=code:server_config]
{
    "server": "https://dart-server:7777",
    "client_key": "000"
}
\end{lstlisting}

The device file must contain a list of client device configurations, each with required key-value pairs \emph{ipAddress}, \emph{port} and \emph{hardware\_config}. In test mode, these can be set to dummy values and the \emph{hardware\_config} can be set to null. A minimal examples a device configuraion file is shown in Listing \ref{code:dummy_device_config}.

\begin{lstlisting}[language=json, caption=Minimal example of a device configuration file with two clients., label=code:dummy_device_config]
{
    "client1": { "ipAdress": "client",
      "port": 2883,
      "hardware_config": null
    },
    
   "client2": { "ipAdress": "client",
      "port": 2883,
      "hardware_config": null
    }
}
\end{lstlisting}

Once the server has been created, it needs to be initialized with the FACT model and some FL stopping criterion by passing them as parameters to the \texttt{initialization\_by\_model} method. The stopping criterion can be selected from various stopping criteria available with FACT, or a custom criterion. For example, the simplest available is the \texttt{FixedRoundFLStoppingCriterion}, which defines a fixed maximum number of rounds for which federated training should be run.

After initialization, the server's \texttt{learning} method should be called to start the training. Behind the scenes, this method makes a call to Fed-DART to asynchronously start the training task on each of the required client devices, sending the model to each one. The server then waits until each client has completed local training and has sent back its updated weights. It then aggregates these and starts another round of training on each client, continuing the loop iteratively until the specified stopping criterion is reached.

When training has completed further optional steps, such as saving the trained model which is available in the Server object, or performing some evaluation, can be included in the main server script as required.
\subsection{Client-side implementation}
\subsubsection{Data importer} On the client side, existing data loading and pre-processing code can be used almost as is by creating a concrete subclass of the \texttt{AbstractDataImporter}. The existing code should then be used to implement the predefined \texttt{load\_data}, \texttt{preprocess\_data} and \texttt{split\_data\_into\_train\_and\_test} abstract methods. 
\subsubsection{Client main script} 
For the main client script, a number of predefined functions should be implemented, that will be called in order by FACT during training. These are the \texttt{init}, \texttt{learn} and, optionally, the \texttt{evaluate} function. These functions should be annotated with \texttt{@feddart}.

The \texttt{init} function receives the \texttt{model\_config}, \texttt{model\_hyperparameters} and \texttt{model\_type} as parameters and should use these to initialize the specified model. The learn function takes \texttt{task\_parameters} and \texttt{global\_model\_parameters} and must use these to update its local model. In the simplest case a straight replacement of local parameters with \texttt{global\_model\_parameters} is sufficient. It is also advisable to save the client parameters during each round. The \texttt{evaluate} function, if implemented, can perform local evaluation of the trained model, saving results as required.

\clearpage
\end{document}